\pdfoutput=1

\documentclass[11pt]{article}

\usepackage[preprint]{acl}

\usepackage{times}
\usepackage{latexsym}

\usepackage[T1]{fontenc}

\usepackage[utf8]{inputenc}

\usepackage{microtype}

\usepackage{inconsolata}

\usepackage{graphicx}

\usepackage{amsmath,amsfonts,bm}

\def\eqref#1{equation~\ref{#1}}

\def\1{\bm{1}}

\DeclareMathAlphabet{\mathsfit}{\encodingdefault}{\sfdefault}{m}{sl}
\SetMathAlphabet{\mathsfit}{bold}{\encodingdefault}{\sfdefault}{bx}{n}

\def\gE{{\mathcal{E}}}

\def\gG{{\mathcal{G}}}

\def\gL{{\mathcal{L}}}

\def\gR{{\mathcal{R}}}

\usepackage{xspace}
\usepackage{adjustbox}
\usepackage{booktabs}
\usepackage{multirow}
\usepackage{amsmath}
\usepackage{gensymb}
\usepackage{soul}
\usepackage{times}    
\usepackage{amssymb}
\usepackage{wrapfig}
\usepackage{floatrow}
\usepackage{rotating}
\usepackage{subcaption}
\usepackage{xcolor}
\usepackage{caption}
\usepackage{tcolorbox}

\usepackage{algorithm}
\usepackage{algorithmic}

\newtcolorbox{promptbox}[1][]{
    top=2pt,
    bottom=2pt,
    fontupper=\small
}

\newcommand{\eg}{\emph{e.g}\xspace}

\newcommand{\myparagraph}[1]{\noindent\textbf{#1}}

\newcommand{\modelshort}{\texttt{RAR}\xspace}
\newcommand{\reasonfull}{Reasoning Chain\xspace}
\newcommand{\reasonfulls}{Reasoning Chains\xspace}

\newcommand{\alignfull}{Knowledge Path\xspace}
\newcommand{\alignfulls}{Knowledge Paths\xspace}

\newcommand{\contextfull}{Graph-aware Reasoning Chain\xspace}
\newcommand{\contextfulls}{Graph-aware Reasoning Chains\xspace}
\newcommand{\unified}{ReAligner\xspace}

\title{Reason-Align-Respond: Aligning LLM Reasoning\\ with Knowledge Graphs for KGQA}

\author{Xiangqing Shen, Fanfan Wang, and Rui Xia\thanks{Corresponding author.} \\
        School of Computer Science and Engineering, \\ Nanjing University of Science and Technology, China \\
        \texttt{\{xiangqing.shen, ffwang, rxia\}@njust.edu.cn}}

\begin{document}

\maketitle

\begin{abstract}
LLMs have demonstrated remarkable capabilities in complex reasoning tasks, yet they often suffer from hallucinations and lack reliable factual grounding. Meanwhile, knowledge graphs (KGs) provide structured factual knowledge but lack the flexible reasoning abilities of LLMs. In this paper, we present Reason-Align-Respond (\modelshort), a novel framework that systematically integrates LLM reasoning with knowledge graphs for KGQA. 
Our approach consists of three key components: a Reasoner that generates human-like reasoning chains, an Aligner that maps these chains to valid KG paths, and a Responser that synthesizes the final answer. 
We formulate this process as a probabilistic model and optimize it using the Expectation-Maximization algorithm, which iteratively refines the reasoning chains and knowledge paths. 
Extensive experiments on multiple benchmarks demonstrate the effectiveness of \modelshort, achieving state-of-the-art performance with Hit scores of 93.3\% and 91.0\% on WebQSP and CWQ respectively. 
Human evaluation confirms that \modelshort generates high-quality, interpretable reasoning chains well-aligned with KG paths. 
Furthermore, \modelshort exhibits strong zero-shot generalization capabilities and maintains computational efficiency during inference. 
\end{abstract}

\section{Introduction}
\label{sec:introduction}

Large language models (LLMs) have exhibited impressive capabilities across a range of complex tasks~\citep{hadi2023survey}, yet their reasoning processes often lack reliable factual knowledge.
This shortcoming reduces interpretability, leads to hallucinations, and causes factual or logical errors~\citep{huang2025survey}.
Knowledge graphs (KGs)~\citep{DBLP:conf/sigmod/BollackerEPST08}, which organize factual knowledge in a structured format, offer strong interpretability and expressive power, making them a reliable source of factual support for LLM reasoning. 
Integrating KGs with LLMs in question answering, \eg, knowledge graph question answering (KGQA), has gained interest as an effective strategy to mitigate hallucinations and enhance interpretability.

\begin{figure}[t]
    \centering
    \includegraphics[width=1\linewidth]{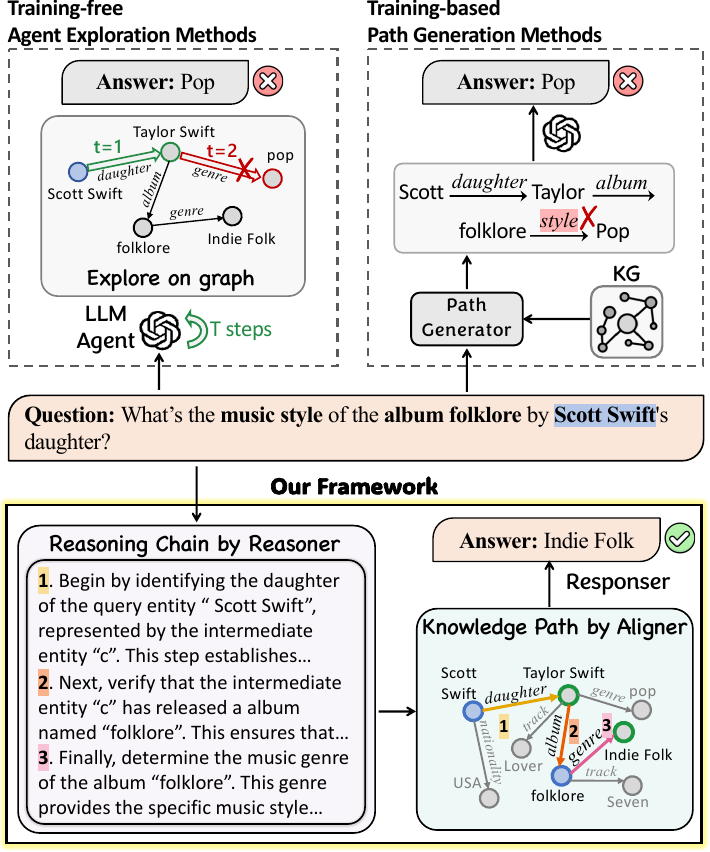}
    \caption{The comparison between our Reason-Align-Respond framework and the existing methods for LLM-based KGQA.}
    \label{fig:motivation}
\end{figure}

The existing LLM-based KGQA studies broadly fall into two main categories: Training-free Agent Exploration methods~\citep{sunthink} and Training-based Path Generation methods~\citep{DBLP:conf/iclr/LuoLHP24}.
The former uses LLMs as agents to explore nodes in KGs, while the latter trains generators to retrieve and generate knowledge paths.
However, both methods lack the holistic reasoning process that humans typically employ when answering questions. This limitation, combined with the semantic gap between natural language descriptions and structured knowledge graphs, often results in reasoning paths that do not align with human logic, lack semantic relevance, or contain unnecessary noise, as shown in Fig.~\ref{fig:motivation}.

On the other hand, the latest deep reasoning LLMs, such as OpenAI-o1~\citep{DBLP:journals/corr/abs-2409-18486} and DeepSeek-R1~\citep{guo2025deepseek}, produce holistic reasoning chains before answering challenging questions, showcasing stronger logical reasoning abilities.
But the reasoning chains obtained through reinforcement learning tend to be verbose, sometimes contain logical fallacies, and amplify hallucinations~\citep{hallucination}.
While our work is not specifically aimed at improving Deepseek-R1,
we believe that using knowledge paths from KGs as constraints for deep reasoning might potentially alleviate these issues.

The two aforementioned aspects are complementary. 
For one thing, allowing LLMs to first perform human-like reasoning in KGQA, can establish a structured, goal-oriented framework that helps reduce invalid or noisy KG path exploration.
For another, while free reasoning by LLMs tends to increase hallucinations, incorporating knowledge paths from KGs provides constraints that help ensure alignment with the knowledge graph, thereby reducing hallucinations.

In this work, we introduce Reason-Align-Respond (\modelshort{}), a novel framework that systematically integrates three modules---Reasoner, Aligner, and Responser---to align LLM reasoning with knowledge graphs for KGQA.
Each of the three modules is a fine-tuned LLM.
Firstly, Reasoner conducts global reasoning for the question, simulating human-like thinking process to generate a reasoning chain that guides LLM's exploration on the KG.
Secondly, Aligner decodes a knowledge path on the KG based on the above reasoning chain. 
Each decoding step ensures correspondence to an actual knowledge triple in the graph.
Finally, Responser leverages both the reasoning chain and the knowledge path to generate the final answer.

We treat both reasoning chain $z_r$ and knowledge path $z_p$ as latent variables,
use a probabilistic model to formalize the distribution of answer $a$ given the question $q$ and KG $\gG$, and propose an end-to-end training algorithm to jointly fine-tune the parameters across all three modules.
Specifically, we employ the expectation-maximization (EM) algorithm to iteratively optimize model parameters while inferring the latent variables. The E-step estimates the posterior probability of latent variables ($z_r$, $z_p$) given observations ($q$, $\gG$, $a$) and samples high-quality reasoning chain and aligned knowledge paths accordingly. The M-step maximizes the evidence lower bound (ELBO) of the log-likelihood to update model parameters.
Through iterative optimization, 
the model progressively refines the reasoning chain and knowledge path, 
and in turn promotes the generation of high-quality responses, 
ultimately forming a stable closed loop.

We conduct extensive evaluation across multiple benchmarks, including WebQSP, CWQ, CommonsenseQA (CSQA), and MedQA, using Freebase, ConceptNet and a medical KG as knowledge graphs. 
The results demonstrate the effectiveness of \modelshort in  improving KGQA performance. 
Compared to 19 baselines across three categories, \modelshort achieves state-of-the-art results. 
Specifically, it achieves Hit scores of 93.3\% on WebQSP and 91.0\% on CWQ, with corresponding F1 score of 87.7\% and 84.8\%, significantly surpassing the existing methods.
Additionally, \modelshort demonstrates strong zero-shot generalizability on CSQA and MedQA.
The EM algorithm demonstrates gradual performance improvements with increasing iterations and shows stable convergence after 200 steps.
Human evaluation of reasoning chains shows that \modelshort generates high-quality, interpretable chains aligned with KG paths, ensuring accurate and reliable reasoning.
We also report the results of \modelshort under different LLM backbone models,
and conduct the ablation study that confirms the importance of each component.
Notably, \modelshort achieves these effective performance while maintaining computational efficiency during inference.
These comprehensive results demonstrate that our approach successfully bridges the gap between LLM reasoning and structured knowledge, offering a promising direction for developing more reliable and interpretable question answering systems.

\begin{figure*}[t]
  \centering
   \includegraphics[width=\textwidth]{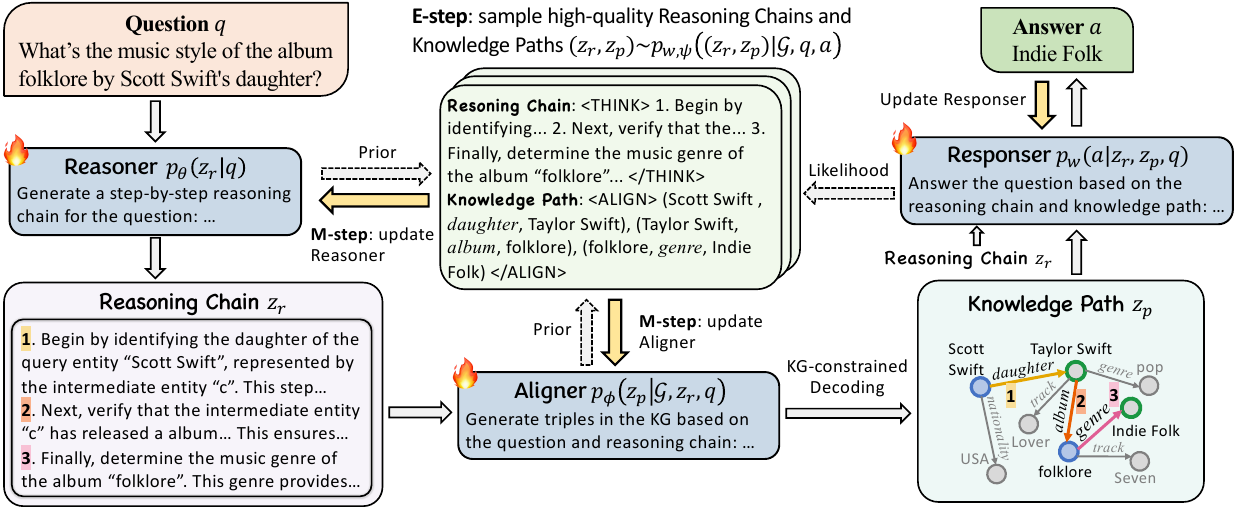}
   \caption{Illustration of our \modelshort framework comprising Reasoner, Aligner, Responser with iterative EM optimization.}
   \label{fig:framework}
\end{figure*}

\section{Approach}
\label{sec:approach}

Knowledge graphs (KGs) contain extensive factual knowledge in the form of triples:
$\gG=\{(e,r,e')|e,e'\in\gE,r\in\gR\}$,
where $\gE$ and $\gR$ denote sets of entities and relations, respectively. 
The goal of knowledge graph question answering (KGQA) is to retrieve relevant facts from a KG $\gG$ and generate an answer $a$ in response to a given natural language question $q$.

\subsection{Task Formalization}
\label{sec:task_formalization}
We introduce two latent variables, a reasoning chain $z_r$ and a knowledge path $z_p$, working together to answer the question $q$ based on a KG $\gG$:
\begin{itemize}
    \item \reasonfull $z_r$ denotes a chain of discrete reasoning steps expressed in natural language, working together to address the question $q$.
    \item \alignfull $z_p$ denotes an interconnected path of knowledge triples extracted from a KG $\gG$.
\end{itemize}

Since neither $z_r$ nor $z_p$ are explicitly annotated in existing KGQA benchmarks, we treat them as latent variables,
and employ a probabilistic model to formalize the distribution of the answer $a$ conditioned on the question $q$ and KG $\gG$ as:
\begin{equation}
\begin{aligned}
\label{equ:main}
&p_{w,\phi,\theta}(a|\gG,q)= \\
&\sum_{z_r,z_p}p_{w}(a|z_r,z_p,q)p_{\phi}(z_p|\gG,z_r,q)p_{\theta}(z_r|q),
\end{aligned}
\end{equation}
where we assume $a$ is conditionally independent of $\gG$ given $(z_r,z_p,q)$, allowing factorization and summation over these conditional probabilities.
On this basis, we introduce our framework, Reason-Align-Respond (RAR), that integrates three modules—Reasoner, Aligner, and Responser—to align LLM reasoning with knowledge graphs for KGQA, as illustrated in Fig.~\ref{fig:framework}.

It is worth noting that each of the three modules is a fine-tuned LLM, with parameters denoted as $\theta$, $\phi$, and $\omega$, respectively. They each utilize the Prompts shown in App.~\ref{app:prompts} to generate reasoning chains, knowledge paths, and answers. 

Firstly, \myparagraph{Reasoner $p_{\theta}(z_r|q)$}
generates a latent reasoning chain $z_r$ to address the question $q$.
The Reasoner generates a reasoning chain in the following format:
\begin{equation}\nonumber
    z_r=\texttt{<THINK>}s_1\dots s_t\texttt{</THINK>},
\end{equation}
where \texttt{<THINK>} and \texttt{</THINK>} are special tokens denoting the start and end of the reasoning process, and each $s_i$ is a discrete reasoning step.

Secondly, \myparagraph{Aligner $p_{\phi}(z_p|\gG,z_r,q)$}
takes the question $q$ and the reasoning chain $z_r$ as inputs to explore KG $\gG$, producing a latent reasoning path $z_p$ that aligns with $z_r$.
Aligner processes the prompt with $q$ and $z_r$ to generate a knowledge path in the following format:
\begin{equation}\nonumber
z_p=\texttt{<ALIGN>}\pi_1\dots\pi_t\texttt{</ALIGN>},
\end{equation}
where \texttt{<ALIGN>} and \texttt{</ALIGN>} mark the beginning and end of the knowledge path.
Each $\pi_i$ is a triple from the KG $\gG$ formatted as $\pi_i=\texttt{<TRI>}(e_i^{h},r_i,e_i^{t})\texttt{</TRI>}$,
where \texttt{<TRI>} and \texttt{</TRI>} bound the triple.

Finally, \myparagraph{Responser $p_{w}(a|z_r,z_p,q)$} generates the final answer $a$ by synthesizing the question $q$, reasoning chain $z_r$, and knowledge path $z_p$.

\subsection{Optimization via the EM Algorithm}
As we have mentioned, each of the three modules is a fine-tuned LLM. In this section, we propose an end-to-end training algorithm to jointly optimize the parameters across all three modules. 
The training objective is the likelihood of the distribution $p_{w,\phi,\theta}(a|\gG,q)$
shown in Eq.~(\ref{equ:main}).
Since our model involves latent variables $z_r$ and $z_p$, we adopt the Expectation-Maximization (EM) algorithm—a principled approach for maximum likelihood estimation (MLE) in latent-variable models~\cite{dempster1977maximum,DBLP:conf/aaai/SenCRG22,DBLP:conf/iclr/QuCXBT21}.

\myparagraph{Unifying Reasoner and Aligner.}
In practice, we unify Reasoner and Aligner by merging their latent variables $z_r$ and $z_p$ into a single one $z=(z_r,z_p)$.
This results in a consolidated module, referred to as \textbf{\unified}, whose parameters denoted by $\psi$. 
Hence, instead of generating $z_r$ and $z_p$ separately, \unified simultaneously outputs both a \reasonfull and a \alignfull, treating it as a single instruction-tuning task with a prompt template (see App.~\ref{app:prompts}).
With this simplification, the conditional distribution for the final answer $a$ is:
\begin{equation}
\label{equ:obj_practice}
    p_{w,\psi}(a|\gG,q) =\sum_{z} p_w(a|q,z)p_\psi(z|\gG,q),
\end{equation}
where $\gG$ denotes the KG, $q$ the question, and $z$ the unified latent variable (combining the \reasonfull and \alignfull).

\myparagraph{Learning Objective.}
We aim to learn the parameters $(w, \psi)$ by maximizing the log-likelihood of the training data with respect to Eq.~(\ref{equ:obj_practice}), written as:
\begin{equation}
\begin{aligned}
\label{eq:obj}
\max_{w,\psi}\mathcal{O}(w,\psi) =\log\mathbb{E}_{z\sim p_\psi(z|\gG,q)}[p_w(a|q,z)].
\end{aligned}
\end{equation}
According to Jensen's inequality, we have:
\begin{equation}
    \label{equ:elbo}
    \mathcal{O}(w,\psi) \geq \underbrace{
    \mathbb{E}_{q(z)}\log ({\frac 
 {p_w(a|q,z)p_\psi(z|\gG,q)}{q(z)})
 } 
 }_{\gL_\text{ELBO}},
\end{equation}
where $q(z)$ is a variational distribution.
Equality holds when $q(z)=p_{w,\psi}(z|\gG,q,a)$, the true posterior of $z$.
The term $\gL_\text{ELBO}$ is the Evidence Lower BOund (ELBO), and maximizing $\gL_\text{ELBO}$ indirectly maximizes $\mathcal{O}(w,\psi)$.

\myparagraph{EM Algorithm.}
The EM algorithm alternates between an E-step and an M-step until convergence:

\subparagraph{E-step.}
Given current parameters $(w^{(t)},\psi^{(t)})$ at iteration $t$, 
E-step updates the variational distribution $q^{(t)}(z)$ by minimizing $\mathrm{KL}(q(z)||p_{w^{(t)},\psi^{(t)}}(z|\gG,q,a))$.
The solution is the posterior of $z$ under the current parameters:
\begin{equation}
\label{equ:sol_posterior}
q^{(t)}(z)=p_{w^{(t)},\psi^{(t)}}(z|\gG,q,a).
\end{equation}

\subparagraph{M-step.}
Keeping $q^{(t)}(z)$ fixed, M-step maximizes $\gL_\text{ELBO}$ in Eq.~(\ref{equ:elbo}) with respect to $w$ and $\psi$.
Ignoring terms that do not depend on $(w,\psi)$, the objective reduces to:
\begin{equation}
\label{eq:M_step}
\begin{aligned}
& Q(w,\psi|w^{(t)},\psi^{(t)}) \\
&=\sum_{(\gG,q,a)} \sum_{z} q^{(t)}(z)\,\log[p_w(a|q,z)p_\psi(z|\gG,q)] \\
& =\underbrace{\sum_{(\gG,q,a)}\sum_{z}q^{(t)}(z)\log p_w(a|q,z)}_{Q_\text{Responser}(w)} \\
& \quad + \underbrace{\sum_{(\gG,q,a)}\sum_{z}q^{(t)}(z)\log p_\psi(z|\gG,q)}_{Q_\text{\unified}(\psi)},
\end{aligned}
\end{equation}
which naturally divides into the instruction-tuning objective for Responser and \unified in Eq.~(\ref{equ:obj_practice}).

By iterative optimization with the EM algorithm, our framework progressively refines its understanding of the question.
This iterative process gradually corrects any flaws in \reasonfulls and \alignfulls, leading to answers that are both higher in quality and more interpretable, while significantly reducing the risk of hallucination.

The workflow of the EM algorithm is shown in Alg.~\ref{alg:optim}, with more details in practice in App.~\ref{app:em}.
\begin{algorithm}[!ht]
    \caption{The EM algorithm in \modelshort{}}
    \label{alg:optim}
    \begin{algorithmic}
        \WHILE{not converge}
            \STATE For each instance, sample $N$ \reasonfulls and \alignfulls $z_I$ from ReAligner $p_\psi$.
            \STATE For each instance, update Responser $p_w$ with $Q_\text{Responser}(w)$ in Eq.~(\ref{eq:M_step}) using $z_I$.
            \STATE \emph{$\boxdot$ E-step:}
            \STATE For each instance, identify $K$ high-quality \reasonfulls and \alignfulls $z_I^h$ from $z_I$ based on Eq.~(\ref{equ:sol_posterior}).
            \STATE \emph{$\boxdot$ M-step:}
            \STATE For each instance, update ReAligner $p_\psi$ according to $Q_\text{\unified}(\psi)$ in Eq.~(\ref{eq:M_step}).
        \ENDWHILE
\end{algorithmic}
\end{algorithm}

\subsection{Techniques During Inference}
During inference, given $q$, Reasoner generates $z_r$, while Aligner produces $z_p$.
Responser synthesizes them to produce $a$.
To enhance performance, we introduce three additional key techniques.

\begin{table*}[t]
    \centering
    \caption{Performance comparison with different baselines on the two KGQA datasets.}
    \label{tab:kgqa}
    \begin{adjustbox}{max width=1.\columnwidth}
        \begin{tabular}{@{}c|l|cc|cc@{}}
            \toprule
            \multirow{2}{*}{Types}           & \multicolumn{1}{c|}{\multirow{2}{*}{Methods}} & \multicolumn{2}{c}{WebQSP} & \multicolumn{2}{c}{CWQ}                                 \\ \cmidrule(l){3-6}
                                             & \multicolumn{1}{c|}{}                         & Hit                        & F1                      & Hit           & F1            \\ \midrule
            \multirow{10}{*}{LLM Reasoning}  & Qwen2-7B \citep{qwen2}                        & 50.8                       & 35.5                    & 25.3          & 21.6          \\
                                             & Llama-2-7B \citep{touvron2023llama}           & 56.4                       & 36.5                    & 28.4          & 21.4          \\
                                             & Llama-3.1-8B \citep{llama3}                   & 55.5                       & 34.8                    & 28.1          & 22.4          \\
                                             & GPT-4o-mini  \citep{gpt4o}                    & 63.8                       & 40.5                    & 63.8          & 40.5          \\
                                             & ChatGPT \citep{chatgpt}                       & 59.3                       & 43.5                    & 34.7          & 30.2          \\
                                             & ChatGPT+Few-shot \citep{brown2020language}    & 68.5                       & 38.1                    & 38.5          & 28.0              \\
                                             & ChatGPT+CoT \citep{wei2022chain}              & 73.5                       & 38.5                    & 47.5          & 31.0              \\
                                             & ChatGPT+Self-Consistency  \citep{wangself}    & 83.5                       & 63.4                    & 56.0          & 48.1              \\ \midrule
            \multirow{4}{*}{Graph Reasoning} & GraftNet  \citep{sun2018open}                 & 66.7                       & 62.4                    & 36.8          & 32.7          \\
                                             & NSM \citep{he2021improving}                   & 68.7                       & 62.8                    & 47.6          & 42.4          \\
                                             & SR+NSM  \citep{zhang2022subgraph}             & 68.9                       & 64.1                    & 50.2          & 47.1          \\
                                             & ReaRev \citep{mavromatis2022rearev}           & 76.4                       & 70.9                    & 52.9          & 47.8          \\ \midrule
            \multirow{10}{*}{KG+LLM}         & KD-CoT \citep{wang2023knowledge}              & 68.6                       & 52.5                    & 55.7          & -             \\
                                             & EWEK-QA \citep{dehghan-etal-2024-ewek}        & 71.3                       & -                       & 52.5          & -             \\
                                             & ToG (GPT-4)  \citep{sunthink}                   & 82.6                       & -                       & 68.5          & -          \\
                                             & EffiQA \citep{dong2024effiqa}                 & 82.9                       & -                       & 69.5          &               \\
                                             & RoG (Llama-2-7B) \citep{luo2024rog}                        & 85.7                       & 70.8                    & 62.6          & 56.2          \\
                                             & GNN-RAG+RA \citep{mavromatis2024gnn}          & 90.7                       & 73.5                    & 68.7          & 60.4          \\ 
                                             & \texttt{GCR} (Llama-3.1-8B + GPT-4o-mini)~\citep{DBLP:journals/corr/abs-2410-13080}                       & 92.2                       & 74.1           & 75.8 & 61.7 \\ 
                                             \cmidrule(l){2-6}
                                             & \modelshort (Llama-3.1-8B + GPT-4o-mini)                            & \textbf{93.3}                       & \textbf{87.7}           & \textbf{91.0} & \textbf{84.8}          \\    \bottomrule
        \end{tabular}%
    \end{adjustbox}
\end{table*}

\myparagraph{KG-constrained Decoding.}
KG-constrained Decoding aims to prevent hallucinated triples that do not exist in the KG.
When generating the \alignfull, Aligner may inadvertently produce triples absent from the KG.
To address this, KG-constrained Decoding restricts the output tokens so that only tokens forming valid KG triples can be produced.
In this way, the generated \alignfull strictly aligns with actual entities and relations in the KG.
Related work~\citep{DBLP:journals/corr/abs-2410-13080,li2024decoding} also attempts to mitigate similar issues; our approach is tailored specifically to our framework.

\myparagraph{\alignfull Expansion.}
\alignfull Expansion addresses the potential incompleteness of initially generated \alignfulls.
To illustrate, consider a question about countries that share borders with the United States.
a \alignfull
{
\setlength{\abovedisplayskip}{1pt}
\setlength{\belowdisplayskip}{1pt}
\begin{equation}\nonumber
\texttt{\small <ALIGN><TRI>(US,borders,Mexico)</TRI></ALIGN>}
\end{equation}
}
\hspace{-5.5pt} is generated by Aligner. 
While correct, this represents only one instance of a broader pattern.
By abstracting the specific instance into a template:
{
\setlength{\abovedisplayskip}{1pt}
\setlength{\belowdisplayskip}{1pt}
\begin{equation}\nonumber
\texttt{\small <ALIGN><TRI>(US,borders,?country)</TRI></ALIGN>},
\end{equation}
}
\hspace{-5.5pt} where $?country$ is a variable, we capture the fundamental relationship structure.
Applied to the KG, this template retrieves all valid instances, such as:
{
\setlength{\abovedisplayskip}{1pt}
\setlength{\belowdisplayskip}{1pt}
\begin{equation}\nonumber
\texttt{\small <ALIGN><TRI>(US,borders,Canada)</TRI></ALIGN>}.
\end{equation}
}
\hspace{-5.5pt} This method transforms a single \alignfull into a comprehensive query template, enabling more complete and exhaustive answers.

\myparagraph{LLM-driven Consolidation.}
LLM-driven Consolidation addresses the challenge of inconsistencies and noise that emerge when sampling multiple \reasonfulls and \alignfulls.
Multiple sampling helps increase coverage and improve the likelihood of correct answers, but inevitably introduces noise and conflicts between samples.
To address this challenge, we propose using a powerful LLM as a ``Consolidator'' that analyzes and integrates multiple \reasonfulls and \alignfulls to derive final answers, following the prompt template detailed in App.~\ref{app:prompts}. 
This approach effectively preserves the benefits of multiple sampling while leveraging the LLM's analytical capabilities to produce reliable answers.

\section{Experiment}
\label{sec:experiment}

\subsection{Experiment Settings}

\myparagraph{Datasets.}
Following previous research~\citep{DBLP:journals/corr/abs-2410-13080, sunthink}, we evaluate our model on three datasets: WebQuestionSP (WebQSP))~\citep{DBLP:conf/acl/YihRMCS16}, Complex WebQuestions (CWQ)~\citep{talmor2018web}, and CommonsenseQA (CSQA)~\citep{talmor2019commonsenseqa}.
The first two datasets use Freebase~\citep{DBLP:conf/sigmod/BollackerEPST08} as the KG, while CSQA leverages ConceptNet~\citep{speer2017conceptnet}, allowing us to assess model generalizability across the unseen KG.

\myparagraph{Baselines.}
We compare \modelshort with 19 baselines across three categories:
LLM reasoning methods, graph reasoning methods, and KG-enhanced LLM reasoning methods.

\myparagraph{Evaluation Metrics.}
For WebQSP and CWQ, we adopt Hit and F1 as evaluation metrics.
Hit checks whether the generated predictions match any correct answer, while F1 evaluates overall answer coverage by balancing precision and recall.
For CSQA, a multiple-choice QA dataset, we use accuracy as the evaluation metric.

\myparagraph{Implementations.}
Our implementation uses Llama-3.1-8B~\citep{llama3} as the backbone for Reasoner, Aligner, and Responser.
To enhance question decomposition, we pretrain both Reasoner and Aligner using 2,000 exemplars demonstrating step-by-step KG-based problem-solving.
For each component, we generate top-$10$ candidates using KG-constrained Decoding and Knowledge Path Expansion, with GPT-4o-mini handling LLM-driven Consolidation.
Details are provided in App.~\ref{app:experiment}.

\subsection{Main Results}
Tab.~\ref{tab:kgqa} shows that \modelshort achieves significant gains on both WebQSP and CWQ datasets, improving the state-of-the-art by 13.6\% and 23.1\% in F1, and by 1.1\% and 15.2\% in Hit respectively.
These remarkable improvements demonstrate the effectiveness of integrating structured KG knowledge with LLM reasoning.

\begin{table}[t]
\caption{Impact of different components on CWQ.}
\begin{adjustbox}{max width=1.\columnwidth}
\begin{tabular}{@{}l|ccc@{}}
\toprule
Variants                                           & F1            & Precision     & Recall        \\ \midrule
\modelshort                         & \textbf{84.8} & \textbf{84.9} & 89.0          \\ \midrule
\modelshort w/o LC & 83.0          & 81.8          & 91.2          \\
\quad w/o Reasoner                  & 80.3          & 75.6          & \textbf{94.7} \\
\quad w/o KD                        & 48.9          & 54.3          & 50.8          \\
\quad w/o KPE                       & 71.7          & 80.2          & 71.7          \\ \bottomrule
\end{tabular}
\label{tab:ablation}
\end{adjustbox}
\end{table}
\subsection{Ablation Study}
As shown in Tab.~\ref{tab:ablation}, removing LLM-driven Consolidation (LC) lowers precision but increases recall, since LC aims to eliminate noisy predictions.
Excluding Reasoner causes a pronounced drop in precision but a rise in recall, indicating that \reasonfulls guide Aligner to explore the KG more accurately, reducing hallucination and noise.
Disabling Knowledge Path Expansion (KPE) diminishes performance, confirming its role in enriching \alignfulls.
Most importantly, removing KG-constrained Decoding (KD) yields the largest performance decrease, underscoring the importance of restricting generation to valid KG paths.

\subsection{Further Analyses}

\begin{figure}[t]
    \centering
    \includegraphics[width=0.8\textwidth]{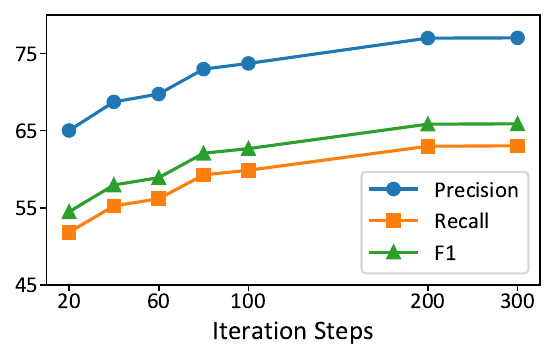}
    \caption{Impact of iteration steps of the EM algorithm.}
    \label{figure:iteration}
\end{figure}
\myparagraph{Impact of Iteration Steps.}
As shown in Fig.~\ref{figure:iteration}, \modelshort exhibits consistent improvement across all metrics during EM updates.
The performance rises rapidly in early stages and continues to refine over iterations, eventually reaching convergence with minimal fluctuations.

\begin{figure}[t]
    \centering
    \includegraphics[width=0.75\textwidth]{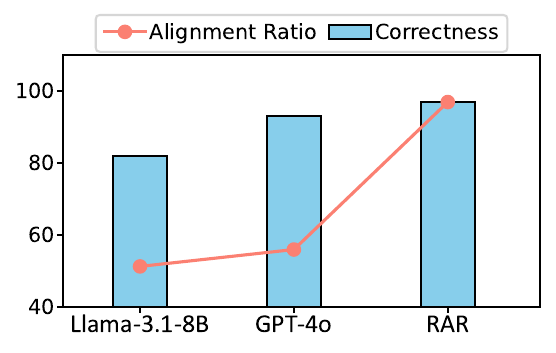}
    \caption{Human evaluation of reasoning chains on CWQ.}
    \label{figure:reasoning_chain}
\end{figure}
\begin{table}[t]
\label{table:efficiency}
\caption{Efficiency and performance of \modelshort compared to different methods on CWQ.}
\begin{adjustbox}{max width=1.\columnwidth}
\begin{tabular}{@{}l|l|c|c@{}}
\toprule
Types                            & Methods                    & Hit           & Avg. Runtime (s) \\ \midrule
\multirow{3}{*}{Path Generation} & GNN-RAG                    & 66.8          & 1.73             \\
                                 & RoG                        & 62.6          & 2.68             \\
                                 & \texttt{GCR}  & 75.8          & \textbf{3.72}    \\ \midrule
\multirow{2}{*}{Agent Exploration}     & ToG                        & 68.5          & 18.89            \\
                                 & EffiQA                     & 69.5          & -                \\ \midrule
Ours                             & \modelshort & \textbf{91.0} & 4.38             \\ \bottomrule
\end{tabular}
\end{adjustbox}
\end{table}
\myparagraph{Quality of \reasonfulls.}
Through manual evaluation of 500 randomly selected samples, we assess the quality of \reasonfulls using two criteria: reasoning correctness and KG alignment.
The correctness metric evaluates whether the \reasonfull successfully solves the given question, while the alignment metric measures how well the reasoning steps correspond to valid KG paths.
As shown in Fig.~\ref{figure:reasoning_chain}, \modelshort substantially outperforms both GPT-4o and baseline methods across both metrics.
These results demonstrate that by aligning \reasonfulls to KG structures, our approach not only improves the reliability of the reasoning process but also enhances its interpretability.

\begin{figure}[t]
    \centering
    \includegraphics[width=0.9\textwidth]{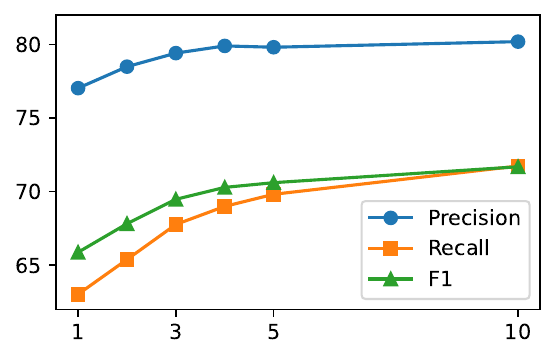}
    \caption{Impact of different beam size on CWQ.}
    \label{figure:beam_size}
\end{figure}
\myparagraph{KG-constrained Decoding Effectiveness.}
We examine the effectiveness of KG-constrained Decoding in mitigating hallucinations and maintaining efficiency.
Our method achieves zero hallucinations in \alignfulls when answers are correct, while without constraints, even correct answers show a 44\% hallucination rate. The efficiency evaluation reveals minimal computational overhead. Particularly noteworthy is the comparison with \texttt{GCR}, the previous state-of-the-art method using KG constraints.
Our approach achieves a 15.2\% improvement in answer accuracy over \texttt{GCR}, with only a marginal increase in runtime from \reasonfull generation.
This modest overhead is well justified by the significant gains in answer reliability and interpretability.

\begin{table}[t]
\caption{Impact of using different LLM backbones for Reasoner, Aligner and LLM-driven Consolidation.}
\begin{adjustbox}{max width=1.\columnwidth}
\begin{tabular}{llcc}
\hline
Components                                                                                                & Variants                           & Hit         & F1            \\ \hline
\multicolumn{1}{l|}{\multirow{5}{*}{\begin{tabular}[c]{@{}l@{}}Reasoner and\\ Aligner\end{tabular}}} & \multicolumn{1}{l|}{Llama-2-7B}    & 84.0          & 68.9          \\
\multicolumn{1}{l|}{}                                                                                     & \multicolumn{1}{l|}{Llama-3.1-8B}  & \textbf{85.2} & \textbf{72.6} \\ \cline{2-4} 
\multicolumn{1}{l|}{}                                                                                     & \multicolumn{1}{l|}{Qwen2-0.5B}    & 71.3          & 56.0          \\
\multicolumn{1}{l|}{}                                                                                     & \multicolumn{1}{l|}{Qwen2-1.5B}    & 72.0          & 56.6          \\
\multicolumn{1}{l|}{}                                                                                     & \multicolumn{1}{l|}{Qwen2-7B}      & 81.4          & 67.1          \\ \hline\hline
\multicolumn{1}{l|}{\multirow{5}{*}{\begin{tabular}[c]{@{}l@{}}LLM-driven\\ Consolidation\end{tabular}}}  & \multicolumn{1}{l|}{GPT-4o-mini}   & 91.0          & 84.8          \\
\multicolumn{1}{l|}{}                                                                                     & \multicolumn{1}{l|}{GPT-4o}        & \textbf{92.8} & \textbf{84.9} \\ \cline{2-4} 
\multicolumn{1}{l|}{}                                                                                     & \multicolumn{1}{l|}{Qwen2-7B}      & 88.7          & 82.2          \\ \cline{2-4} 
\multicolumn{1}{l|}{}                                                                                     & \multicolumn{1}{l|}{Llama-3.1-8B}  & 90.6          & 83.7          \\
\multicolumn{1}{l|}{}                                                                                     & \multicolumn{1}{l|}{Llama-3.1-70B} & 92.4          & 83.0          \\ \hline
\end{tabular}
\label{tab:backbones}
\end{adjustbox}
\end{table}

\myparagraph{Impact of Beam Size.}
As shown in Fig.~\ref{figure:beam_size}, increasing beam size allows \modelshort to explore more potential \reasonfulls and \alignfulls, leading to improved performance across all metrics.
This demonstrates that examining multiple candidate solutions helps identify better \reasonfulls and \alignfulls for responses of higher quality.

\myparagraph{Impact of different LLM Backbone.}
Tab.~\ref{tab:backbones} demonstrates that larger LLMs generally achieve better performance, with Llama-3.1-8B and GPT-4o delivering the strongest results for backbone and LLM-based Consolidation (LC), respectively.

\myparagraph{Zero-shot Generalizability to Unseen KGs.}
\begin{table}[t]
\label{tab:unseen}
\caption{Zero-shot transferability to unseen KG.}
\begin{adjustbox}{max width=1.\columnwidth}
\begin{tabular}{@{}lcc@{}}
\toprule
Model       & CSQA        & MedQA       \\ \midrule
GPT-4o-mini & 91          & 75          \\
\texttt{GCR}         & 94          & 79          \\
\modelshort         & \textbf{94} & \textbf{80} \\ \bottomrule
\end{tabular}
\end{adjustbox}
\end{table}
Following \citet{DBLP:journals/corr/abs-2410-13080}, we evaluate \modelshort's zero-shot transfer capabilities on CSQA~\citep{talmor2019commonsenseqa} and MedQA~\citep{jin2021disease}.
The results show that \modelshort achieves superior zero-shot performance compared to GPT-4o-mini on both datasets.
Notably, \modelshort achieves comparable performance to \texttt{GCR}, as both methods leverage the KG to constrain the decoding process to enhance reasoning without requiring additional training on the target KG.

\subsection{Case study}
\begin{figure*}[t]
\centering

\includegraphics[width=1\linewidth]{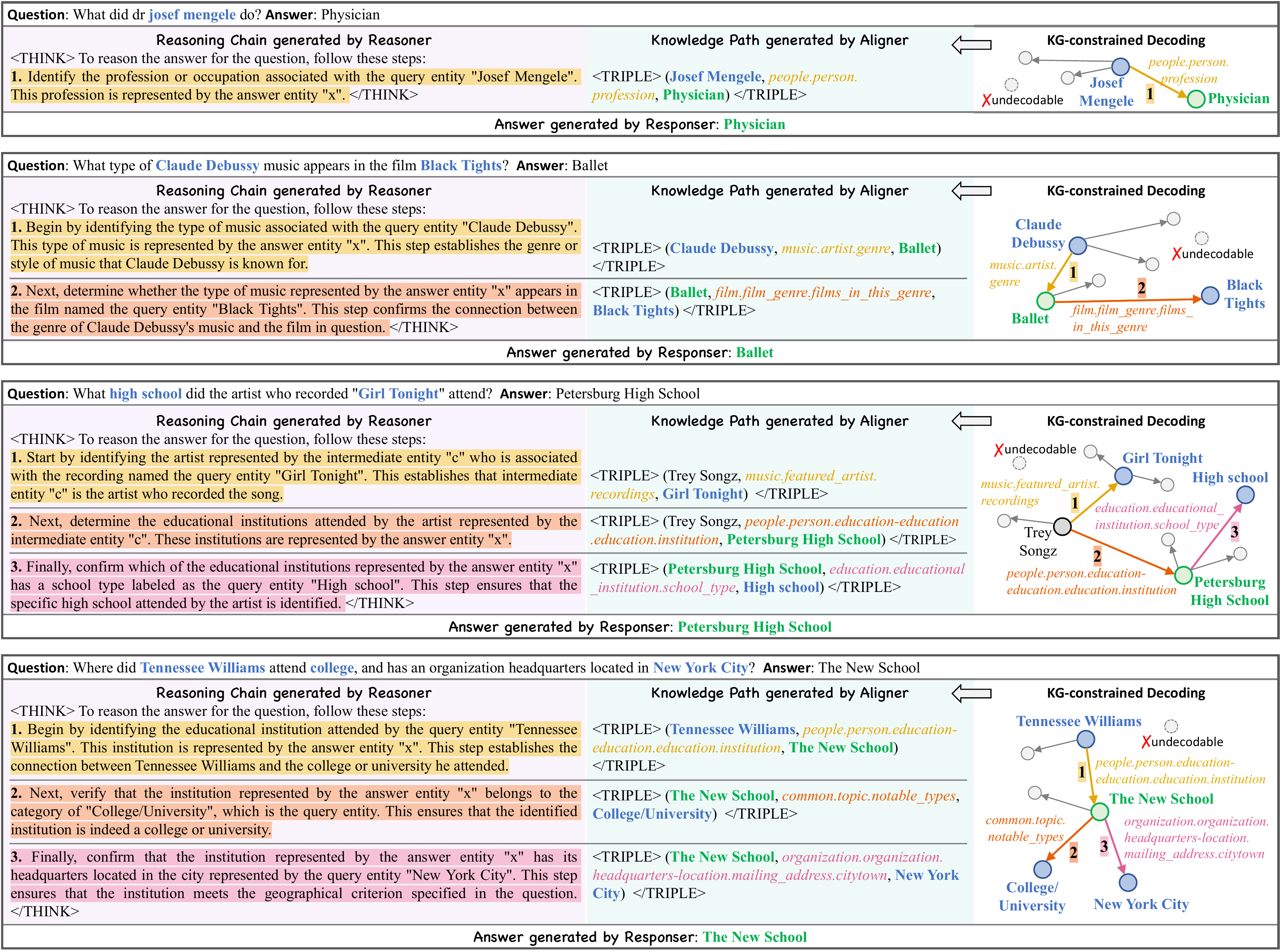}
\caption{Examples of \reasonfulls and \alignfulls generated by \modelshort under different iteration steps.}
\label{fig:case_study}
\end{figure*}
Fig.~\ref{fig:case_study} illustrates the qualitative results by our \modelshort framework.
To investigate the effect of EM iterations, in Fig.~\ref{fig:case_study_step}, we also present two representative cases that demonstrate the evolution of \modelshort's behavior across iterations of the EM algorithm.
These cases showcase distinct improvements in \modelshort's ability to generate \reasonfulls and \alignfulls.
In Case 1, we examine \modelshort's response to the query ``What high school did the artist who recorded 'Girl Tonight' attend?''
The early-stage model demonstrates suboptimal verification behavior by directly searching for high school information without following the proper verification sequence:
first identifying educational institutions, then verifying the specific type as high school.
This \reasonfull prevents proper alignment with the \alignfull in the KG.
In contrast, the later-stage model generates both \reasonfulls and \alignfulls effectively, producing a higher quality \reasonfull that successfully aligns with the \alignfull in the KG.
Case 2 examines \modelshort's handling of ``What type of Claude Debussy music appears in the film Black Tights?''
Here, we observe a different pattern of improvement.
While the early-stage model generates the same \reasonfull as the later-stage model, it fails to generate \alignfulls that fully align with and reflect this reasoning, resulting in a misaligned \alignfull that do not lead to the correct answer.
The later-stage model maintains consistency between \reasonfulls and \alignfulls, thus arriving at the correct answer.
These cases validate the effectiveness of the EM approach.

\section{Related Work}
\label{sec:related_work}

\myparagraph{LLM Reasoning.} 
Recent progress in LLMs have spurred extensive research to improve deep reasoning. 
One line of work focuses on prompting-based methods, which elicit intermediate reasoning steps during inference, such as Chain-of-Thought (CoT) prompting~\citep{DBLP:conf/nips/Wei0SBIXCLZ22}.
Building on CoT, self-consistency~\citep{DBLP:conf/iclr/0002WSLCNCZ23} generates multiple reasoning traces and selects the most consistent answer.
Tree-of-Thought~\citep{DBLP:conf/nips/YaoYZS00N23} explores branching steps in a tree structure to uncover optimal solutions.
Beyond prompting, researchers have investigated fine-tuning LLMs on reasoning tasks~\citep{DBLP:journals/corr/abs-2212-08286,DBLP:conf/nips/HoffmanPDDLPSSV23}, including reinforcement learning methods~\citep{openai_o1, guo2025deepseek} that encourage more complex multi-step reasoning before arriving at a final answer.

\myparagraph{KG-enhanced LLM Reasoning.}
Despite their remarkable performance, LLMs still face limitations such as incomplete or outdated domain knowledge, interpretability challenges, and the potential for hallucinations.
To address these issues, a growing body of work has focused on integrating LLMs with KGs~\citep{pan2023unifying}.
KD-CoT~\citep{wang2023knowledge} enhances CoT by retrieving relevant facts from external KGs, guiding LLMs with more reliable information.
RoG~\citep{luo2024rog} employs a plan–retrieve–reason pipeline that explicitly fetches KG-based reasoning paths to ground the final answer.
GCR~\cite{DBLP:journals/corr/abs-2410-13080} further mitigates hallucinations by enforcing graph-constrained decoding, ensuring that every reasoning step aligns with valid KG connections.
GNN-RAG~\citep{mavromatis2024gnn} leverages a graph neural network for effective KG retrieval, while StructGPT~\citep{jiang2023structgpt} and ToG~\citep{sunthink} treat the LLM as an agent that navigates the KG, assembling multi-hop paths to produce more trustworthy answers.

\section{Conclusion}
In this paper, we present \modelshort, a novel framework that integrates LLM reasoning with knowledge graphs for KGQA through three key components—Reasoner, Aligner, and Responser. We formulate this process as a probabilistic model and optimize it using the Expectation-Maximization algorithm. Through extensive experiments on multiple benchmarks, we demonstrate that \modelshort achieves state-of-the-art performance while maintaining interpretability and computational efficiency. 
These results demonstrate that \modelshort successfully bridges the gap between LLM reasoning and structured knowledge, offering a promising direction for building reliable and interpretable QA systems.

\section*{Limitations}

One limitation of our framework lies in the computational overhead introduced by Reasoner.
In certain cases, especially for complex queries, the \reasonfull generated by Reasoner can become relatively long, increasing resource consumption.
However, our experimental results demonstrate that the performance gains from incorporating \reasonfull justify this additional cost, striking a practical balance between efficiency and effectiveness.
Another limitation concerns the generalizability to specialized domains.
Though our framework, trained on Freebase-based KGQA datasets, shows improved generalization to unseen KGs compared to previous methods, its performance on highly specialized knowledge graphs (\eg, medical KGs) remains to be enhanced.
Improving adaptation to domain-specific KGs presents a promising direction for future research.



\bibliography{custom, mybib}

\newpage

\appendix

\section{Rationale and Details of the EM Algorithm}
\label{app:em}

We provide the motivation and the details of applying the EM algorithm to optimize our framework.

\subsection{Overview}
We have two modules:
\begin{itemize}
    \item \textbf{\unified}, parameterized by $\psi$, which generates \contextfulls $z$ given $(\mathcal{G}, q)$. 
    \item \textbf{Responser}, parameterized by $w$, which predicts the final answer $a$ given the question $q$ and candidate \contextfull $z$.
\end{itemize}

Given a training set of triples $\{(\mathcal{G}, q, a)\}$, our objective is to maximize:
\[
\mathcal{O}(w,\psi) \;=\; 
\sum_{(\mathcal{G},q,a)} \log 
\Bigl( \mathbb{E}_{z \sim p_{\psi}(z \mid \mathcal{G}, q)} 
\bigl[ p_{w}(a \mid q, z) \bigr] \Bigr).
\]
Because exact marginalization over $z$ can be expensive, we employ an EM-style approach to iteratively refine both modules.

\subsection{Rationale}

The selection of the EM algorithm is fundamentally motivated by the central challenge and objective of the \modelshort: generating latent natural language (NL) reasoning steps for KGQA.

\subsubsection{Core Challenge: Generating Latent NL Reasoning}
\begin{itemize}
    \item \modelshort aims to produce complex, human-like NL reasoning chains – the intermediate ``thought process'' connecting a question to its answer using a Knowledge Graph (KG).
    \item Crucially, these detailed reasoning steps are \textbf{not available} in standard KGQA training datasets. They constitute \textbf{latent variables} within our model.
\end{itemize}

\subsubsection{Inadequacy of Direct Supervision Methods}
\begin{itemize}
    \item In standard SFT applied within RoG~\citep{DBLP:conf/iclr/LuoLHP24}, relation sequences absent from the original training data are often generated. To label these sequences for training, the process frequently relies on heuristics to create pseudo-gold labels, such as identifying the shortest KG path between query and answer entities, which can be potentially noisy.
    \item This heuristic-based supervision is insufficient for training models to generate the complex, multi-step, logically nuanced \textbf{NL reasoning} that \modelshort targets, especially when multiple constraints are involved.
\end{itemize}

\subsubsection{EM as the Principled Approach for Latent Variables}
\begin{itemize}
    \item EM is the standard, principled approach for parameter estimation in models with latent variables.
    \item It provides a formal framework to optimize the Reasoner (generating the latent NL chain) and the Aligner (grounding the chain to the KG) \textbf{without requiring explicit supervision} for the intermediate NL reasoning steps.
    \item Optimization is guided indirectly by maximizing the likelihood of observing the correct final answer, conditioned on the feasibility of the generated reasoning chain being aligned to the KG.
\end{itemize}

\subsubsection{Necessity of Iterative Refinement}
\begin{itemize}
    \item Generating coherent, long-form NL reasoning is challenging. Initial attempts, especially early in training, are likely to be imperfect or logically flawed.
    \item The iterative nature of the EM algorithm is well-suited for this progressive refinement: \textbf{E-step} identifies the most likely or ``best'' latent reasoning chains produced by the current model that  successfully link the question to the correct answer via a feasible KG alignment. This step essentially evaluates the current reasoning quality based on outcomes; \textbf{M-step} updates the parameters of the Reasoner and Aligner models by training them on these high-quality reasoning chains identified in the E-step. This step aims to make the models generate more chains similar to the successful ones.
    \item This iterative E-M loop allows the system to gradually improve the quality, logical coherence, and KG-alignability of the generated latent reasoning, as demonstrated qualitatively in Fig.~\ref{figure:reasoning_chain}.
\end{itemize}

\subsubsection{Connections to Reinforcement Learning}
\myparagraph{Connection to Implicit RL.}
The EM algorithm, as applied in \modelshort, can be viewed as a form of implicit Reinforcement Learning:
\begin{itemize}
    \item The \textbf{E-step} acts like a selection or filtering mechanism based on the quality of the reasoning chain, implicitly assigning a high reward (\eg, 1) to successful chains (reaching the correct answer via KG alignment) and low reward (\eg, 0) to unsuccessful ones.
    \item The \textbf{M-step}, particularly the Reasoner update (maximizing $log p(z_hat_I|q)$ for selected high-quality chains $z_hat_I$), mathematically resembles a policy gradient update ($E[R * \nabla \log p(z|q)] \approx R * \nabla \log p(\hat{z}|q)$) where $R$ is effectively this implicit binary reward.
\end{itemize}
Thus, EM reinforces the generation of ``good'' reasoning chains without the need for explicit reward engineering.

\myparagraph{Why EM Was Preferred Over Explicit Reinforcement Learning.}
While the EM process here shares similarities with RL (see next point), we opted for EM over explicit RL formulations (like PPO) for several practical reasons:
\begin{itemize}
    \item Reward Function Design: Crafting a good reward function (`R`) that accurately captures the quality of multi-step \textbf{NL reasoning} is non-trivial.
    \item Training Complexity and Cost: Explicit RL methods often lead to higher computational costs and potentially unstable training.
    \item Efficiency and Simplicity: EM, derived naturally from the maximum likelihood objective for latent variable models, offers a more direct, mathematically grounded, and often simpler optimization pathway for our specific problem structure.
\end{itemize}

\subsection{Details of EM Algorithm}
\subsubsection{Step 1: Updating Responser}

\begin{enumerate}
\item \textbf{Sample Candidate \contextfulls.}
 For each training example $(\mathcal{G}, q, a)$, sample $K$ \contextfulls:
 \[
 z_k \;\sim\; p_{\psi}(z \mid \mathcal{G}, q), 
 \quad k = 1, \dots, K.
 \]
 Let $\hat{z} = \{\,z_1, z_2, \ldots, z_K\}$.

\item \textbf{Approximate the Objective for $w$.}
 The term
 \[
 \log \mathbb{E}_{z \sim p_{\psi}(z \mid \mathcal{G},q)} 
    \bigl[ p_w(a \mid q, z)\bigr]
 \]
 is approximated by
 \[
 \log \biggl(\tfrac{1}{K} \sum_{k=1}^K p_{w}(a \mid q, z_{k}) \biggr).
 \]
 We then take gradients (w.r.t.\ $w$) and update $w$ so that $p_{w}(a \mid q, z)$ is more likely to produce the correct $a$ for the sampled \contextfulls.

\item \textbf{Result.}
 After updating $w$, Responser $p_w(a \mid q, z)$ is better aligned with whatever \contextfulls $p_\psi$ currently emits.
\end{enumerate}

\subsubsection{Step 2: EM-Style Update for \unified}

After $w$ is updated, we refine the \unified $p_{\psi}(z \mid \mathcal{G}, q)$. In EM terms, we view $z$ as a latent variable:

\myparagraph{E-Step (Posterior Inference)}

\begin{itemize}
    \item \textbf{Compute / Re-rank \contextfulls.} Re-sample or re-rank the $K$ \contextfulls using the updated $p_w$. We want \contextfulls that are ``most aligned'' with the correct answer $a$. Formally:
    \[
    p_{w,\psi}(z \mid \mathcal{G}, q, a) \;\propto\; 
    p_w(a \mid q, z)\, p_{\psi}(z \mid \mathcal{G}, q).
    \]
    \item \textbf{Scoring.} For a single \contextfull $z$, define
    \[
    S(z) \;=\; 
    \log p_w(a \mid q,z) 
    \;+\; 
    \log p_\psi(z \mid \mathcal{G}, q).
    \]
    \item \textbf{Selecting High-Quality \contextfulls.} 
    Rank (or sample) \contextfulls by $S(z)$ and select the top set $z_I = \{\text{top-}K \text{ \contextfulls }\}$. 
\end{itemize}

\myparagraph{M-Step (Update $\psi$)}

\begin{itemize}
    \item Treat $z_I$ (the selected high-quality \contextfulls) as if they were observed.
    \item Update $\psi$ by maximizing:
    \[
    \log p_{\psi}(z_I \mid \mathcal{G}, q)
    \;=\;
    \sum_{z\,\in\,z_I}
    \log p_{\psi}(z \mid \mathcal{G}, q)
    \]
    \item In practice, this amounts to standard fine-tuning (e.g., instruction tuning or teacher forcing) of $p_\psi$ on the newly identified high-quality \contextfulls.
\end{itemize}

\subsubsection{Complete Iteration}

A single iteration of our EM-style algorithm proceeds as follows:
\begin{enumerate}
    \item \textbf{(Update $w$):} 
          For each sample $(\mathcal{G}, q, a)$, draw $K$ \contextfulls from $p_\psi$, then update $w$ by maximizing 
          \[
            \log \left(\tfrac{1}{K}\sum_{k=1}^K 
                p_w(a \mid q, z_k)\right).
          \]
    \item \textbf{(E-Step):}
          Using the updated $w$, compute 
          \[
          p_{w,\psi}(z \mid \mathcal{G}, q, a)
          \;\propto\; 
          p_w(a \mid q, z)\, p_{\psi}(z \mid \mathcal{G}, q).
          \]
          Select high-quality \contextfulls $z_I$ from the candidates.
    \item \textbf{(M-Step):} 
          Update $\psi$ by maximizing $\log p_{\psi}(z_I \mid \mathcal{G}, q)$, 
          i.e.\ fine-tune $p_{\psi}$ so that it is more likely to emit $z_I$ in the future.
\end{enumerate}

This loop can be repeated until convergence or for a fixed number of epochs.

\subsubsection{Practical Variations}
\label{app:practical}

\begin{enumerate}
    \item \textbf{Top-$K$ vs.\ Full Posterior.} 
          Instead of summing/sampling over all subsets, it is simpler to pick the top-$K$ \contextfulls by $S(\cdot)$.
    \item \textbf{Skipping Responser Optimization.}
    To further improve efficiency, we can skip optimizing Responser. LLMs often possess strong zero-shot summarization or question-answering capabilities, which means they can produce high-quality answers from given \contextfulls without additional training. As a result, we can treat an LLM as a pre-optimized Responser and focus solely on updating \unified, thereby reducing overall computation.
\end{enumerate}

\section{More Related Work}

\subsection{Comparison with Agent Exploration Methods}

To situate \modelshort within the KGQA landscape, we first contrast it with representative agent exploration methods such as ToG~\citep{sunthink}.  
Although both paradigms comprise stages that can be informally mapped to \emph{reasoning} and \emph{grounding}, their internal principles diverge markedly.

\paragraph{Training methodology and optimization.}
\modelshort is trained with a mathematically grounded expectation–maximisation (EM) procedure that \emph{explicitly and stably} refines two complementary capabilities: the NL \emph{Reasoner} and the KG‑aware \emph{Aligner}.  
By contrast, many agent methods rely more heavily on prompt or workflow engineering~\cite{cheng-etal-2024-call,gu-etal-2024-middleware,huang-etal-2024-queryagent,li-etal-2023-shot,nie2024code}; they seldom perform task‑specific optimization that directly targets the core reasoning mechanism.

\paragraph{Accuracy, reliability, and complexity handling.}
The synergy between NL reasoning, KG‑constrained alignment, and EM‑guided supervised fine‑tuning translates into markedly higher accuracy and robustness for \modelshort (see Tab.~\ref{tab:kgqa}).
Empirically, its explicit decomposition allows it to cope well with multi‑hop and conjunctive constraints that are challenging for purely prompt‑driven agents.

\paragraph{Resource consumption.}
Once training is finished, inference in \modelshort can be carried out by a collection of relatively small, specialized, fine‑tuned models—one each for the Reasoner, Aligner, and Responser.
This modularity yields the efficiency gains reported in Tab.~\ref{table:efficiency}.
Agent systems, in contrast, often incur higher latency and cost because they perform several large‑LLM calls while exploring the KG.

Taken together, these differences show that \modelshort is not a mere variant of the agent paradigm; its EM‑centered optimization strategy and KG‑constrained decoding constitute a distinct design that offers both practical efficiency and stronger empirical performance.

\subsection{Comparison with Path Generation Methods}

\modelshort also differs fundamentally from path generation methods such as RoG~\citep{DBLP:conf/iclr/LuoLHP24} and GCR~\citep{DBLP:journals/corr/abs-2410-13080}.  
Where those systems directly predict a linear sequence of KG relations, ours maintains a higher‑level, human‑readable plan in natural language and lets the Aligner ground \emph{each} step to KG triples.

Specifically, \textbf{(i)} the Reasoner produces a multi‑step NL chain that expresses the conceptual logic;  
\textbf{(ii)} the Aligner incrementally matches every NL step to concrete triples through KG‑constrained decoding; and  
\textbf{(iii)} the Responser integrates evidence from both the NL chain and the aligned KG path to craft the final answer.  
Training again relies on EM—iteratively improving latent NL chains that can be aligned and that ultimately yield correct answers—whereas RoG and related work usually depend on direct SFT with shortest KG paths that may be noisy supervision signals.

\paragraph{Illustrative example.}
For the query \emph{``What did Dr Josef Mengele do?''} the two paradigms unfold differently:

\begin{itemize}
  \item \emph{RoG.}  An LLM planner outputs the relation \texttt{people.person.profession}; a symbolic retriever then follows this edge in the KG to obtain the triple \texttt{(Josef Mengele, profession, Physician)}, and the answer \emph{Physician} is returned.
  \item \emph{\modelshort.}  
        \textbf{Step 1} (A Reasoner step) proposes: ``Identify the profession associated with \emph{Josef Mengele}.''  
        The Aligner grounds this to the same triple as above.  
        The Responser finally reports \emph{Physician}, explicitly citing both the reasoning chain and the grounded path.
\end{itemize}

\paragraph{Handling complex constraints.}
Because RoG represents reasoning as a \emph{single} linear relation sequence, it struggles with conjunctive queries such as \emph{``presidents who were also actors''}; after following \texttt{profession $\rightarrow$ Actor}, it cannot backtrack to verify \texttt{profession $\rightarrow$ President}.  
\modelshort, in contrast, naturally decomposes the query into two successive NL steps (“find presidents” $\rightarrow$ “filter those who are actors”) and grounds each step separately, ensuring both constraints are satisfied.

\paragraph{Robustness to noisy supervision.}
Shortest‑path supervision can be incorrect when more semantically plausible paths exist.  By letting EM discover latent NL chains that are \emph{both} alignable and answer‑bearing, \modelshort avoids this brittleness and achieves the quality gains visualized in Fig.~\ref{figure:reasoning_chain} of the submission.

In summary, the collaboration of an explicit NL Reasoner, a KG‑constrained Aligner, and EM‑based optimization endows \modelshort with a distinctive combination of interpretability, flexibility, and empirical strength that is not achieved by prior agent exploration or path generation methods.

\section{Experimental Setup}
\label{app:experiment}

\subsection{Fine-tuning Datasets and Knowledge Graph}
For evaluation, we use two benchmark KGQA datasets: WebQuestionSP (WebQSP)~\citep{DBLP:conf/acl/YihRMCS16} and Complex WebQuestions (CWQ)~\citep{talmor2018web}.
To ensure fair comparison, we adopt identical train and test splits as previous works~\citep{jiang2022unikgqa,luo2024rog}.
The detailed statistics of these datasets are presented in Tab.~\ref{tab:datasets}.
Both WebQSP and CWQ are based on Freebase~\citep{DBLP:conf/sigmod/BollackerEPST08}.
To reduce computational overhead and memory consumption, we utilize the same subgraphs as previous work~\citep{luo2024rog}. 
Additionally, we preprocess Freebase by converting CVT (Compound Value Type) nodes, which represent n-ary relationships, into binary relationships by concatenating edge labels with ``-'' as the delimiter, following~\citet{li2024decoding}.

\begin{table*}[th]
    \begin{adjustbox}{max width=1.\columnwidth}
    \centering
    \caption{Statistics of datasets.}
    \label{tab:datasets}
    \begin{tabular}{@{}c|cc|cccc@{}}
    \toprule
    \multirow{2}{*}{Dataset} & \multicolumn{2}{c|}{Dataset Statistics} & \multicolumn{4}{c}{Statistics of Answer Numbers}                                \\ \cmidrule(l){2-7} 
                             & \#Train             & \#Test            & \#Ans = 1 & 2 $\geq$ \#Ans $\leq$ 4 & 5 $\geq$ \#Ans $\leq$ 9 & \#Ans $\geq$ 10 \\ \midrule
    WebQSP                   & 2,826               & 1,628             & 51.2\%    & 27.4\%                  & 8.3\%                   & 12.1\%          \\
    CWQ                      & 27,639              & 3,531             & 70.6\%    & 19.4\%                  & 6\%                     & 4\%             \\ \bottomrule
    \end{tabular}%
    \end{adjustbox}
\end{table*}

\subsection{Datasets for Cold Starting}
To initialize the training of Reasoner and Aligner, we leverage high-quality \reasonfulls and \alignfulls derived from SPARQL queries in the WebQSP and CWQ datasets.
This approach prevents the models from generating malformed outputs during early training stages.

The SPARQL queries in these datasets represent the gold-standard \reasonfull that human experts would use to solve questions using the KG.
We decompose these SPARQL queries according to a predefined grammar, breaking them into atomic chunks that each represent a single reasoning step. For each chunk, we query Freebase to obtain the corresponding triples, then use GPT-4o to generate natural language reasoning chains based on these retrieved triples.
Through this process, we generate a dataset of 2,000 high-quality \reasonfulls with their corresponding \alignfulls for each question.
This dataset enables us to perform cold-start pretraining of Reasoner and Aligner, teaching them to generate well-structured, step-by-step \reasonfulls and \alignfulls with appropriate special tokens, a crucial foundation for subsequent optimization using the EM algorithm.

\subsection{Hyperparameters}
For Responser, we adopt Llama-3.1-8B~\citep{llama3} without fine-tuning based on our preliminary experiments (detailed analysis in App.~\ref{app:practical}).
For both Reasoner and Aligner, we conduct extensive experiments with various lightweight LLMs ranging from 0.5B to 8B parameters~\citep{qwen2,touvron2023llama,llama3}.
All models share the same hyperparameter configuration:
training for 3 epochs with a batch size of 4 and a learning rate of 2e-5.
We employ a cosine learning rate scheduler with a warmup ratio of 0.03.
The training is performed on 4 A6000 GPUs for each model variant.

\section{Details of KG-constrained Decoding}
For efficient knowledge graph operations, we implemented a Virtuoso-based Freebase instance with distributed high-speed SPARQL querying capabilities. Our system achieves a throughput of 2,000 requests per second, enabling rapid graph traversal and neighborhood node retrieval during the constrained decoding process. This high-performance infrastructure allows us to efficiently retrieve next-hop candidates for token constraints directly from the knowledge graph.

\section{Case Study of Different Iteration Steps}
\begin{figure*}[t]
\centering

\includegraphics[width=1\linewidth]{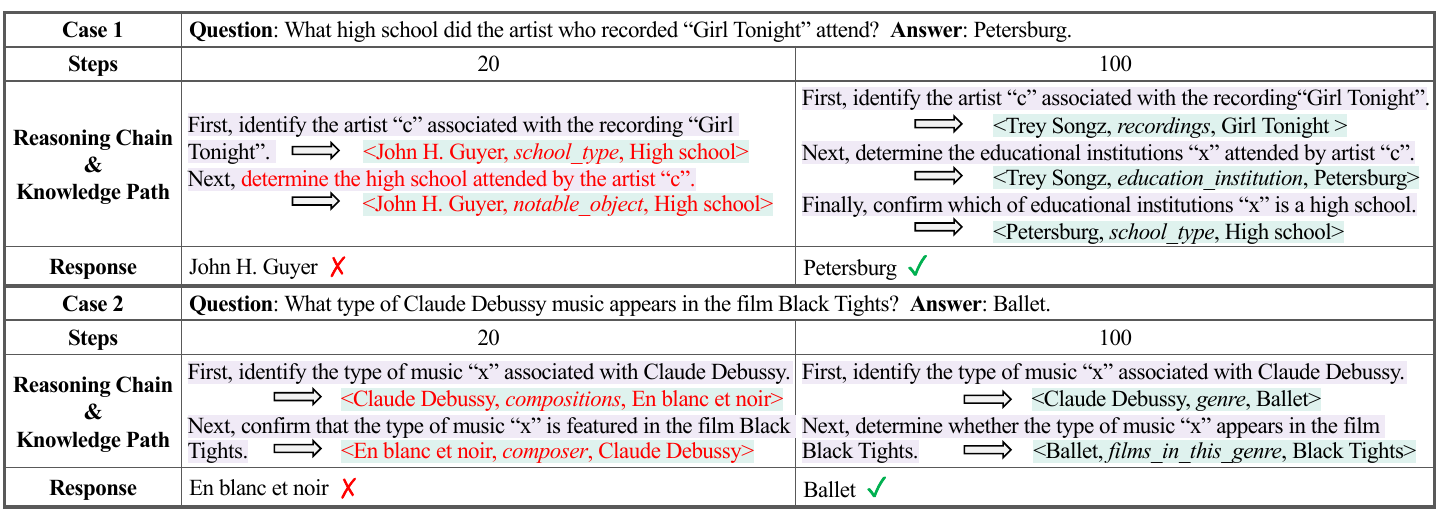}
\caption{Examples of \reasonfulls and \alignfulls generated by \modelshort under different iteration steps.}
\label{fig:case_study_step}
\end{figure*}

In Fig.~\ref{fig:case_study_step}, we provide two examples to investigate the effect of iteration steps of the EM algorithm.

\section{Templates and Prompts}
\label{app:prompts}

In this section, we illustrate all the prompts used in the experiments.

\myparagraph{\reasonfull Template.}
The template of \reasonfulls generated by Reasoner is shown in Fig.~\ref{fig:reasoning_chain_template}, where each $s_i$ is a discrete reasoning step in natural language.
\begin{figure}[h]
    \centering
    \begin{minipage}{0.99\columnwidth}
        \centering
        \begin{tcolorbox}[title=Reasoning Chain Template]
            \small
            \texttt{<THINK>}$s_1 s_2 s_3 \dots s_n$ \texttt{</THINK>}
        \end{tcolorbox}
    \end{minipage}
    \caption{The template for the \reasonfull generated by Reasoner.}
    \label{fig:reasoning_chain_template}
\end{figure}

\myparagraph{\alignfull Template.}
The template of \alignfull generated by Aligner is shown in Fig.~\ref{fig:knowledge_path_template}, where $e$ abd $r$ denotes the entities and relations from the KG.
\begin{figure*}[h]
    \centering
    \begin{minipage}{0.99\columnwidth}
        \centering
        \begin{tcolorbox}[title=Knowledge Path Template]
            \small
            \texttt{<ALIGN>}\texttt{<TRIPLE>} <|> $e_1$ <|> $r_1$ <|> $e_1^{'}$\texttt{</TRIPLE>}\ldots\texttt{<TRIPLE>} <|> $e_n$ <|> $r_n$ <|> $e_n^{'}$\texttt{</TRIPLE>}\texttt{</ALIGN>}
        \end{tcolorbox}
    \end{minipage}
    \caption{The template of \alignfulls generated by Aligner.}
    \label{fig:knowledge_path_template}
\end{figure*}

\myparagraph{\unified Prompt.}
The prompt for instructing \unified is shown in Fig.~\ref{fig:realigner_prompt}, where the task is to generate the \reasonfull and \alignfull given the question and the KG.
\begin{figure*}[htb]
    \centering
    \begin{minipage}{0.99\columnwidth}
        \centering
        \begin{tcolorbox}[title=\unified Prompt]
            \small
            ============================= Prompt Input ================================
    
            Generate a step-by-step thinking process for the given question. Ensure the thinking process is aligned with triples in the knowledge base.
            
            \vspace{10pt}
            
            Question:
            
            \texttt{<Question>}

            \vspace{10pt}
            
            Query entities:
            
            \texttt{<Question Entities>}
    
            \vspace{10pt}
    
            ============================= LLM Output ================================
    
            Thinking process: 
            
            \texttt{<Reasoning Chain>}

            \vspace{10pt}
    
            Align Process: 
            
            \texttt{<Knowledge Path>}
        \end{tcolorbox}
        \vspace{1mm}
    \end{minipage}
    \caption{The prompt template for \unified.}
    \label{fig:realigner_prompt}
\end{figure*}

\myparagraph{Responser Prompt.}
The prompt for instructing Responser is shown in Fig.~\ref{fig:responser_prompt}, where the task is to generate the final answer based on the given the question and the generated \reasonfull and \alignfull.
\begin{figure*}[htb]
    \centering
    \begin{minipage}{0.99\columnwidth}
        \centering
        \begin{tcolorbox}[title=Responser Prompt]
            \small
            ============================= Prompt Input ================================
    
            Generate a step-by-step thinking process for the given question. Ensure the thinking process is aligned with triples in the knowledge base.
            
            \vspace{10pt}
            
            Question:
            
            \texttt{<Question>}

            \vspace{10pt}
            
            Query entities:
            
            \texttt{<Question Entities>}
    
            \vspace{10pt}

            Thinking process: 
            
            \texttt{<Reasoning Chain>}

            \vspace{10pt}
    
            Align Process: 
            
            \texttt{<Knowledge Path>}

            \vspace{10pt}

            Summarization:
    
            ============================= LLM Output ================================

            \texttt{<Answer>}
            
        \end{tcolorbox}
        \vspace{1mm}
    \end{minipage}
    \caption{The prompt template for Responser.}
    \label{fig:responser_prompt}
\end{figure*}

\myparagraph{LLM-driven Consolidation Prompt.}
The prompt for LLM-driven Consolidation is shown in Fig.~\ref{fig:llm_driven_consolidation_prompt}.
We use \modelshort to generate $K$ \alignfulls and hypothesis answers for each question. The \alignfulls and hypothesis answers are provided to general LLMs to answer the questions without fine-tuning.
\begin{figure*}[h]
    \centering
    \begin{minipage}{0.99\columnwidth}
        \centering
        \begin{tcolorbox}[title=LLM-driven Consolidation Prompt]
            \small
            ============================= Prompt Input ================================
    
            Relevant triples:

            \texttt{<Knowledge Path 1>}. Therefore, a possible answer could be: \texttt{<Hypothesis Answer 1>}

            $\ldots$

            \texttt{<Knowledge Path K>}. Therefore, a possible answer could be: \texttt{<Hypothesis Answer K>}
            
            \vspace{10pt}

            Question: 
            
            \texttt{<Question>}

            \vspace{10pt}

            Based on the reasoning paths, please answer the given question. Please keep the answer as simple as possible and only return answers. Please return each answer in a new line.
            
            \vspace{10pt}
            ============================= LLM Output ================================
    
            \texttt{<Answer 1>}

            \texttt{<Answer 2>}

            $\ldots$

        \end{tcolorbox}
        \vspace{1mm}
    \end{minipage}
    \caption{The prompt template for LLM-driven Consolidation.}
    \label{fig:llm_driven_consolidation_prompt}
\end{figure*}

\end{document}